# i-MYO: A Hybrid Prosthetic Hand Control System based on Eye-tracking, Augmented Reality and Myoelectric signal*

Chunyuan Shi, Dapeng Yang, Jingdong Zhao, Li Jiang

*Abstract*— Dexterous prosthetic hands have better grasp performance than traditional ones. However, patients still find it difficult to use these hands without a suitable control system. A new hybrid myoelectric control system, termed *i-MYO*, is presented and evaluated to solve this problem. The core component of the i-MYO is a novel grasp-type switching interface based on eye-tracking and augmented reality (AR), termed i-GSI. With the i-GSI, the user can easily switch a grasp type (six total) for a prosthetic hand by gazing at a GazeButton. The i-GSI is implemented in an AR helmet and is integrated, as an individual module, into the i-MYO system. In the i-MYO system, the myoelectric signal was used to control hand opening/closing proportionally. The operation of the i-MYO was tested on nine healthy subjects who wore HIT-V hand on the forearm and manipulated objects in a reach-and-grasp task. It was also tested on one patient who had an inferior myoelectric signal and was required to control the HIT-V hand to grasp objects. Results showed that in 91.6% of the trials, inexperienced *healthy* subjects accomplished the task within 5.9 s, and most failed trials were caused by a lack of experience in fine grasping. In addition, in about 1.5% of trials, the subjects also successfully transferred the objects but with a non-optimal grasp type. In 97.0% of the trials, the subjects spent ~1.3 s switching the optimal grasp types. A higher success rate in grasp type (99.1%) for the untrained *patient* has been observed thanks to more trials conducted. In 98.7 % of trials, the *patient* only needed another 2 s to control the hand to grasp the object after switching to the optimal grasp type. The tests demonstrate the control capability of the new system in multi-DOF prosthetics, and all inexperienced subjects were able to master the operation of the i-MYO quickly within a few pieces of training and apply it easily.

*Index Terms*— Prosthetic hand, Hybrid control system, Eye-tracking, AR

## I. INTRODUCTION

Prosthesis refers to a device that can help amputees restore hand function and improve their quality of life[1]. However, most commercial prosthetic hands are simple and unable to meet the daily needs of patients [2]. Many patients hope that these hands can have better grasp performance and that fingers can move independently [3]. In research, many multi-fingered dexterous prosthetic hands [4]–[6] meet the functional requirements and can reproduce a variety of grasp types [7] to deal with different grasp tasks. But most users do not accept these hands due to the absence of a suitable control system [8]. A reliable and intuitive control system to support dexterous manipulation is needed for advanced prostheses [9].

The myoelectric control system (MEC) has always been the mainstream for prosthetic hand control [8]. In a MEC, weak EMG signals are obtained from the patient's stump through EMG electrodes and processed to control the prosthesis[9]. The classic proportional control method [9], [10] and the on/off control method were proposed in the 1960s, still being adopted for most commercial prosthetics [11]. These two methods are simple and intuitive but have insufficient functions, which cannot support the multi-grasp type control of dexterous prosthetics [12]. Although the three-amplitude state method [13], [14] and the finite state machine method (FSM) [2], [15] can increase the controllable degrees of freedom (DOF), they are not real-time and have poor intuitiveness and large burden.

Researchers have focused on the EMG pattern recognition algorithm (PR)[11] to improve the MEC system in the past few decades. PR assumes the existence of distinguishable and repeatable patterns among EMG activities[16]. Thus a method based on multi-channel signals (> four channels) and classifiers (such as Support Vector Machine (SVM)[17], [18], Linear Discriminate Analysis (LDA [19])), Artificial Neural Networks (ANN [20], [21]), etc.) can establish the mapping relationship between muscle activity and grasp type. However, this method is slow in response, non-proportional control [11], and is not robust [9] due to electrode crosstalk [22], forearm motion[23], forearm loading [24], and other confounding factors [25].

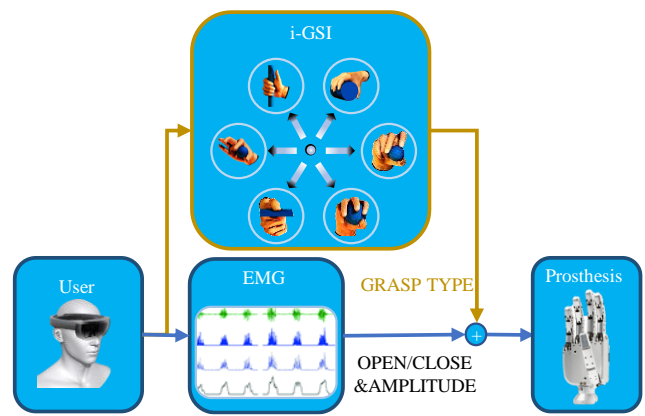

Fig. 1. Schematic diagram of new system principle. i-GSI, a grasp-type switching interface based on augmented reality and eye-tracking. The new system uses i-GSI for switching a grasp type and EMG for controlling hand closing\opening.

* This work is partially supported by NSF Grant #52075114, Interdisciplinary Research Foundation of HIT (IR2021218), and Postdoctoral Scientific Research Development Fun (LBH-W18058) to D.Yang. Corresponding author: Dapeng Yang (yangdapeng@hit.edu.cn).

The authors are from the State Key Laboratory of Robotics and System, Harbin Institute of Technology (HIT), Harbin 150080, China. D. Yang and J. Zhao are also from the Artificial Intelligence Laboratory (HIT), Harbin 150001, China (e-mail: Chunyuan_Shi, yangdapeng, zhaojingdong, jiangli01@hit.edu.cn).

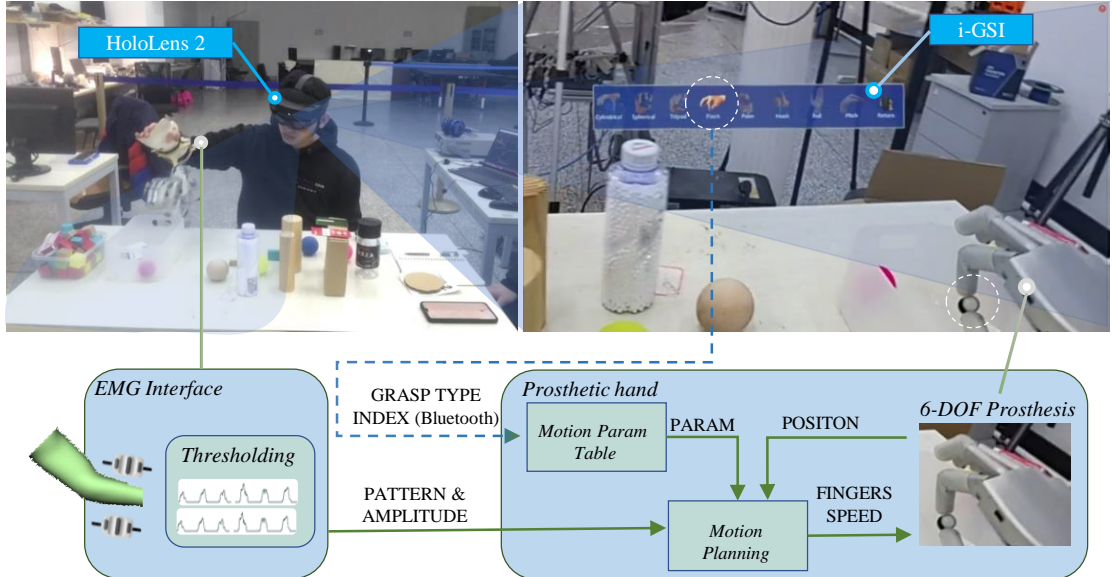

Fig. 2. The i-MYO control system. The first image shows that the subject was pinching a bead using the prosthetic hand attached to his forearm, and the second image is the scene the subject saw through the AR helmet, HoloLens 2.

Moreover, the available channels are limited due to muscle loss resulting from amputation, further reducing availability [26]. Although targeted muscle reinnervation technology (TMR) can alleviate the problem of muscle loss and improve classification accuracy [27]–[29], most patients reject it due to the many risks associated [26]. On the other hand, algorithms based on Linear Regression (LR) [30], Non-Negative Matrix (NMF) [31], and Deep Learning (DL) [32] can regress multi-DOF synchronous proportional control signals from multi-channel EMGs. It is also difficult to operate the hand dexterously because of less DOF (two to three).

It is necessary to introduce other sensors to make up for the deficiency of the EMG-only control system [8]. The EMG hybrid control system (EMG-HCS) [26] has been further developed in recent years. A hybrid control system refers to a system with more than two control signal sources [26], [33]. In an EMG-HCS, EMG is still used to control hand opening and closing (proportionally control), other signals, such as plantar pressure [34], [35], tongue movement signal [36], and radiofrequency tags [37], etc., are used to switch the hand's grasp type. However, current EMG-HCS implementations still have an insufficient performance to switch the grasp type [26]. It is still complicated to use and non-real-time. Besides, combining the features of different signals as the input of the PR algorithm is also a hybrid control method, but most are offline analyses[9], [38]. Artificial vision is a newly developed method in recent years, and many challenges await being solved [26], [39], [40], such as clutter background and target object detection.

The authors have recently proposed a new grasp-type switching interface (i-GSI) based on augmented reality and eye-tracking to solve the grasp-type problem [41]. With the i-GSI, users can easily switch between six grasp types by gazing at a holographic grasp-type icon, and even a novice can master the i-GSI very quickly.

This paper proposes a novel hybrid prosthetic hand control system based on i-GSI and myoelectric signal (i-MYO) (shown in Figure 1), demonstrating how the i-GSI as a standalone module can be integrated into a myoelectric dexterous prosthetic hand [6]. The i-MYO can be implemented on commercial mobile devices. The goal of the current study was to test the operation performance of i-MYO. A total of nine healthy novice subjects participated in the reach-grasp experiment, of which the results refer to the efficacy of grasping the objects. Further, the i-MYO was tested in one patient with a congenital upper limb deficiency. The result of the patient proves that even a subject with an extremely poor EMG signal can operate the i-MYO well.

## II. MATERIALS AND METHODS

### A. i-MYO Contol System

The i-MYO control system includes a grasp-type switching interface (i-GSI), a two-site myoelectric interface (MYO), and a multi-DOF prosthetic hand, as shown in Figure 2. For a target object, the user can use the i-GSI to select an optimal grasp type and then use the MYO to control the prosthetic hand to grasp or release the object in a proportional way.

*1）i-GSI:* the i-GSI is responsible for switching grasp types and mainly includes a hologram panel with nine GazeButtons. The GazeButton refers to a button with advanced gaze-based interactions [42]. The hologram panel is directly superimposed in the user's vision, and the GazeButton can be triggered when the user gazes at it. This gaze behavior means that the human observes a position for more than 120 ms, and the threshold parameter of gaze time in this study is set at 200 ms for stable control. Of the nine buttons, six are grasp types: *Cylindrical*, *Spherical*, *Tripod*, *Pinch*, *Lateral*, and *Hook;* these are the basic grasp types for manipulating daily life object*s* [43] and are usually used as indicators for designing dexterous prosthetic hands. A grasp-type index will be sent to the prosthetic hand through Bluetooth when the relative GazeButton is triggered.

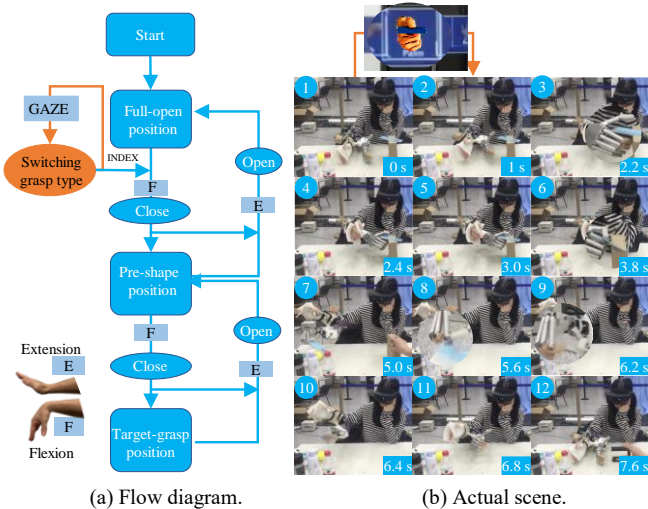

(a) Flow diagram.  (b) Actual scene.
Fig. 3. The operation flow of the i-MYO system. The user was switching the garget grasp type in the first two images. The hand in image ③ was at the pre-shape position, the one in image ⑥ was at the target-grasp position, and the image ⑧ shows that the hand released the object.

The other three buttons are reserved for other functions, such as wrist control, and will not be discussed in this paper. The i-GSI is implemented in an untethered self-contained holographic device (Holoens2, Microsoft, USA) with an eye tracker. Our previous work provides more detailed information about the hologram panel and the gaze algorithm[41].

2) *Myoelectric Interface (MYO):* MYO is used to get the user's intention to proportionally control the prosthetic hand opening\closing from a two-site myoelectric signal. The two-site signal was recorded using two OttoBock dry electrodes (Otto Bock, Germany, 13E200 = 50) over a pair of antagonist muscles (flexor and extensor). The two-site EMG inputs have a sufficient signal ratio thanks to the electrode's preprocessing and were directly thresholded with individually adjustable levels to generate opening/closing control commands. The amplitude of the signal is mapped to the scale factor of the maximum finger speed (90°/s). The MYO's algorithm was implemented on a controller connected with the two-site electrodes, and the controller was housed inside a prosthetic palm.

3) *Prosthetic hand:* The prosthetic hand is used to manipulate daily life objects and emulated by a robot hand, HIT-V [6], attached to the subject's forearm by a 3D-printed bypass. The HIT-V has five anthropomorphic fingers actuated by six DC motors: five motors for finger flexion/extension and one for thumb adduction/abduction. All degrees of freedom contain position sensors, thus realizing position control. A mapping table with motion parameters of six grasp types was saved offline in the controller housed inside the palm. These parameters, such as the pre-shape position of fingers, were obtained by trial and test. Another parameter, the grasp-type index, is obtained from the i-GSI via Bluetooth. The index was used to configure the hand by the look-up table and could only change when the hand was between the full-open and pre-shape positions (as detailed in Section II.B). Yet another parameter, the maximum speed scale factor, was set and mapped linearly proportional to the EMG amplitude.

A personal host was used to adjust system parameters (i.e., EMG threshold, gaze time threshold) for different users during the experiment, which is unnecessary for clinical applications.

*B. Operation Flow of the i-MYO System*

The operation of the i-MYO system is effortless (shown in Figure 3). Firstly, the user gazes at the target GazeButton to switch the grasp type when the fingers are spread out. Secondly, the user approaches the prosthetic hand to the object and elicits an EMG signal (flexing the wrist) to control the prosthetic hand from a full-open position to a pre-shape position. Then, the user controls the hand to continue to close and grasp the object by observing the scene. After transporting the object successfully to another position, the user elicits another EMG signal (stretching the wrist) to open the prosthetic hand to release the object.

### III. EXPERIMENTS

*A. Subjects & Sensor Calibration*

One patient and nine healthy subjects were recruited to test the operation of the i-MYO system. Two healthy subjects had a few experiences with EMG control, and others had none (see more in TABLE I). Evaluating the i-MYO system from the patient with an extremely poor EMG signal can better show the system's applicability to patients. The patient has a congenital upper limb deficiency, and the stump is less than one-fifth of the forearm. Due to the lack of exercise, the arm was as thin as that of a child, with little strength. He did not have the right-hand muscle memory such as flexing/extending the wrist and did not have phantom limb sensation[44]. Therefore, his EMG signal was far inferior to those acquired from the patients with radial amputation. The EMG electrodes were placed over two muscles he could independently activate repeatedly, and the muscles were near the end of the elbow.

For different users, the magnifications and thresholds of the EMG electrodes were adjusted respectively to obtain a good signal-to-noise ratio. Subjects calibrated the eye tracker using an APP of the HoloLens 2 system. The position of the holographic panel was fine-tuned to make the user comfortable,

TABLE I
SUBJECT INFORMATION

| Subject | Handed-ness | Gender | Age | Height (CM) | Weight (KG) | BMI | Prosthetic control experience |
|---|---|---|---|---|---|---|---|
| 1 | R | Male | 24 | 170 | 59 | 20.42 | ☆☆☆☆☆ |
| 2 | R | Male | 24 | 173 | 63 | 21.05 | ☆☆☆☆☆ |
| 3 | R | Female | 28 | 162 | 49 | 18.67 | ☆☆☆☆☆ |
| 4 | R | Male | 26 | 170 | 76 | 26.30 | ☆☆☆☆☆ |
| 5 | R | Male | 24 | 178 | 78 | 24.62 | ☆☆☆☆☆ |
| 6 | R | Male | 29 | 170 | 61 | 21.11 | ★★☆☆☆ |
| 8 | R | Male | 28 | 170 | 75 | 25.95 | ★☆☆☆☆ |
| 7 | R | Male | 31 | 170 | 66 | 22.84 | ☆☆☆☆☆ |
| 9 | R | Male | 27 | 175 | 73 | 23.84 | ☆☆☆☆☆ |
| 10* | L | **Male** | **26** | **178** | **62** | **19.57** | ★★★★★ |

* Amputee
All experiments were approved by the university's Ethical Committee (NO.HIT-2021009) and conformed to the Declaration of Helsinki.

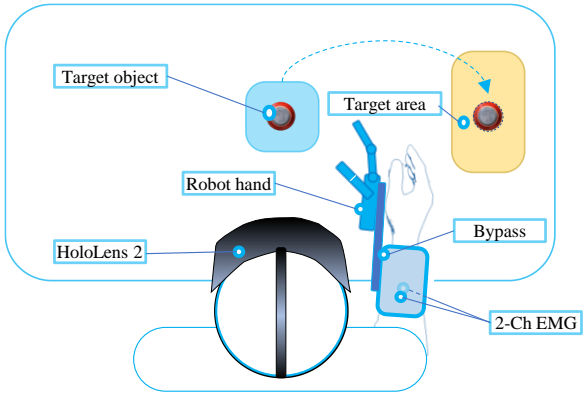

Fig. 4. Experimental setup for healthy subjects. With the robot hand attached to his right forearm by a bypass, the subject sits on the chair comfortably and wears the AR helmet on his head. The two EMG electrodes are also put over the right forearm. He needs to manipulate the robot hand to transfer the object in front of him to the target area at his right.

and the threshold parameter of the gaze (as described in Section II.A) was fine-tuned to adapt to user habits. The prosthetic hand was attached to healthy subjects' right forearms by a 3D-printed bypass.

### B. Experimental Protocols & Data Analysis

*1) Test Task:* The healthy subjects were required to complete a reach-and-grasp task to test the operation of the i-MYO. This task aims to test if the i-MYO can help the user restore typical daily activities. As shown in Figure 4, the subject in this task needed to manipulate the robot hand attached to his forearm to reach, pick up, transport, and place the target object in the target area on the table. As for the patient, his right arm is too thin and powerless to wear the robot hand (> 600 g); thus, the robot hand was fixed on the fixture. He was required to switch a target grasp type and trigger the EMG signal to control the hand to hold the object. This task for the patient aimed to test if a subject with extremely poor EMG could also activate a myoelectric signal to control the prosthetic hand to grasp and release different objects.

Twenty-one household objects listed in Table II were selected as the target objects, including one sample for *Hook* and four samples for every other grasp type. Each healthy subject needed to carry out eight blocks of 24 trials, and the patient needed twenty blocks to gather sufficient data for statistical analysis. The object for *Hook* would appear four times in a block. The objects for *Pinch* were small (i.e., the bead, S$\Phi$1.5 cm) and would be a challenge for the subjects.

*2) Experiment Procedures:* Before experimenting on each subject, the assistant calibrated the sensors and introduced the task to the subject. The subject would take less than ten minutes to practice this system. During practice, the user would try to manipulate the hand to grasp the bead or mini padlock five times to learn how to use *Pinch.* After practice, the user had a ten-minute break before the test. During the test, the assistant randomly placed an object of which the grasp type was non-repeating in each trial. The user observed the object simultaneously so as not to trigger a GazeButton subconsciously. After being instructed, the user started to manipulate. Once finished a block of twenty-four trials, the user rested for five minutes to prevent forearm muscle fatigue.

*3) Metrics:* Six metrics listed in TABLE III were defined and used to evaluate the performance of the i-MYO system. **1)** Grasp-type success rate (SR-G, for *all* subjects): the grasp type was deemed correct if the *final grasp type* used to grasp the object was optimal according to classification in TABLE III. **2)** Time spent switching the *correct* grasp type (T-G, for *all* subjects). **3)** Task success rate for the *healthy* subjects (SR-HT): a trial was considered successful if the healthy subject used the *correct* grasp type to accomplish the task in which the subject was required to pick the object up and place it at the target area. Otherwise, the trial would be considered a failure once an *incorrect* grasp type was used, even if the subject accomplished the task (e.g., using *Pinch* to transfer an object that *Tripod* should have transferred). **4)** The time a *healthy* subject spends on one trial (T-HT). **5)** The task success rate for the *patient* (SR-PT): a trial was considered successful if the *patient* elicits an EMG signal to control the robot hand grasping the objects with the *correct* grasp type. **6)** The time the *patient* spends on one trial (T-PT). The time criteria for T-G, T-HT, and T-PT are illustrated in TABLE III.

TABLE II
TARGET OBJECTS

| No. | Grasp Type | Object | Size (cm) | IMG |
|---|---|---|---|---|
| 1 | Cylindrical | Powder bottle | 15 × Φ6.3 | |
| 2 | | Plastic bottle 1 | 18 × Φ5.2 | |
| 3 | | Plastic bottle 2 | 17.8 × Φ5.8 | |
| 4 | | Sauce bottle | 13.5 × Φ6 | |
| 5 | Spherical | Toy ball 1 | SΦ6.3 | |
| 6 | | Toy ball 2 | SΦ6.3 | |
| 7 | | Toy ball 3 | SΦ6.3 | |
| 8 | | Toy ball 4 | SΦ6.3 | |
| 9 | Tripod | Toy brick | 5.9 × 2.8 × 1.4 | |
| 10 | | Plug | 3.7 × 3.2 × 2 | |
| 11 | | Medicine pack box | 5.5 × 3.3 × 3.3 | |
| 12 | | Small Yarn Ball | SΦ4.4 | |
| 13 | Pinch | Small wooden cube | 1.8 × 1.8 × 1.8 | |
| 14 | | Mini eraser | 2.8 × 1.7 × 0.9 | |
| 15 | | Mini padlock | 5 × 3 × 1.9 | |
| 16 | | Bead | SΦ1.5 | |
| 17 | Lateral | Transdermal patch | 13 × 11 × 0.05 | |
| 18 | | Plastic card | 8.5 × 5.3 × 0.1 | |
| 19 | | Tea bag | 10 × 4.7 × 0.1 | |
| 20 | | Ruler | 16 × 4 × 0.2 | |
| 21 | Hook* | Plastic handle | 9.5 × 12.5 Φ3 | |

* Printed 3D model, shaped like a doorknob, will appear four times in each block test.

TABLE III
METRICS WITH DEFINITION

| Metrics | Definition | Schematic diagram |
|---|---|---|
| SR-G | Grasp-type success rate | |
| T-G | The time spent switching *correct* grasp type. | |
| SR-HT | The task success rate for the *healthy* subjects. | |
| T-HT | The time a *healthy* subject spends on one trial. | |
| SR-PT | The task success rate for the *patient*. | |
| T-PT | The time *patient* spends on one trial. | |

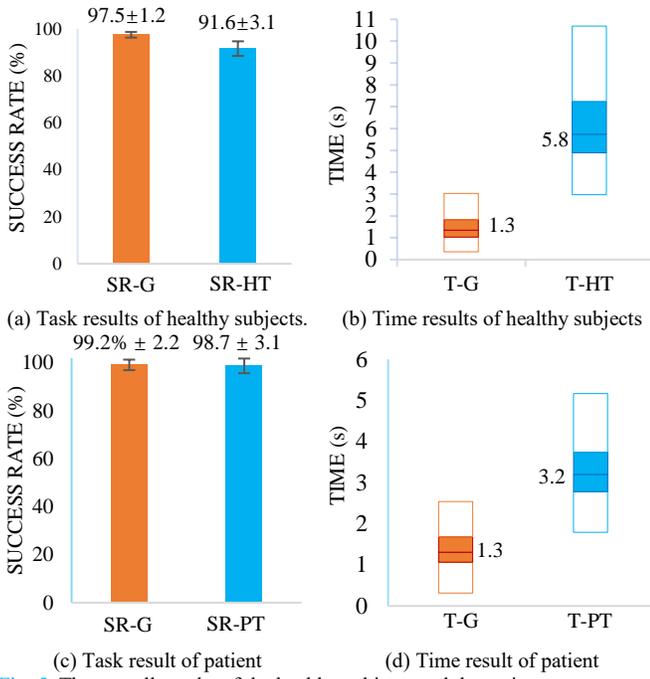

Fig. 5. The overall results of the healthy subjects and the patient.

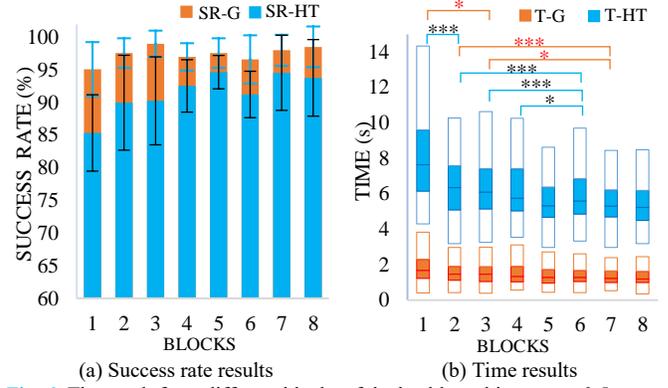

Fig. 6. The result from different blocks of the healthy subjects. $p_* < 0.5$, $p_{**} < 0.01$, $p_{***} < 0.001$.

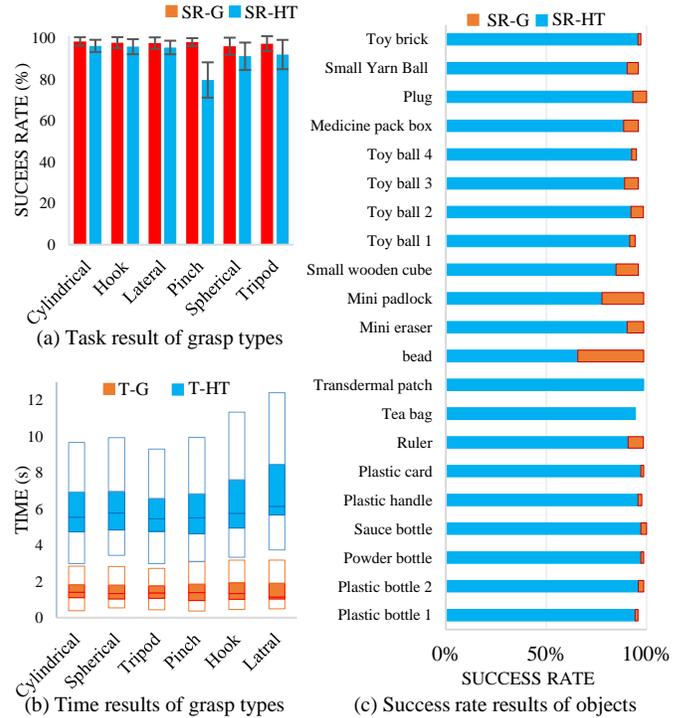

Fig. 7. Results of different grasp types or objects

*4) Data Analysis Methods:* this study used the box plots with median, inter-quartile range, and extremum to represent the time data collected in this paper since it did not fit a normal distribution (Kolmogorov–Smirnov normality test). A Wilcoxon signed-rank test was used to determine the significance of the paired data between two groups, and a Friedman test and the Bonferroni adjustment were used when the paired data was from more than two groups.

## IV. RESULTS & ANALYSIS

Nine healthy subjects finished a total of 1728 trials (nine subjects, eight 24-trial blocks), and the patient finished 480 trials (one subject, 20 24-trial blocks). Overall, the healthy subjects with no prior training successfully transferred objects in 91.6% of the cases and switched the optimal grasp types in 97.5%, as shown in Figure 5(a). It means that only one-third of failures were attributed to i-GSI. Furthermore, in about 1.5% of cases, the subjects successfully transferred the objects but with a non-optimal grasp type one that the subject considered optimal (e.g., using *Tripod* to grasp objects that *Pinch* should have grasped), mainly in early trials. These subjects spent about 5.8 s transferring an object and about 1.3 s switching the grasp type, as shown in Figure 5(b). Figure 5(c) showed that the patient could successfully trigger the EMG to control the hand holding the objects in 98.7% of trials in around 3.2 s and spend around 1.3 s to switch the optimal grasp types in 99.2% of trials.

The analysis of different blocks showed that the healthy subjects with no prior training could quickly master the operation of the i-MYO system, and the operation of i-GSI was natural and stable. Most unsuccessful tasks happened in the first three blocks, especially the first block, with only 85% of cases successful, as shown in Figure 6(a). But the figure also clearly showed a significant improvement from the first to the fifth block (85% to 95%), and Figure 6(b) showed the time spent reduced from 7.6 s to 5.0 s. In addition, no significant difference was observed in the fifth to eighth block. The switch success rate is 95% in the first block and is increased to 99% in the third block. The time spent switching the grasp type decreased from 1.7 s in the first block to 1.2 s in the eighth block, and no significant difference was observed from the fourth block to the eighth block.

The grasping success rate according to different grasp types is shown in Figure 7. Figure 7(a) shows that objects grasped by *Pinch* (80% success rate) were the most difficult to transfer since these objects are too small to be precisely grasped (i.e., the bead: $S\Phi 1.5 \text{ cm}^3$, mini padlock: $5.0 \times 0.3 \times 1.9$ cm). The robot hand fingers often closed on empty air in the first few blocks since the subjects failed to predict the contact position of

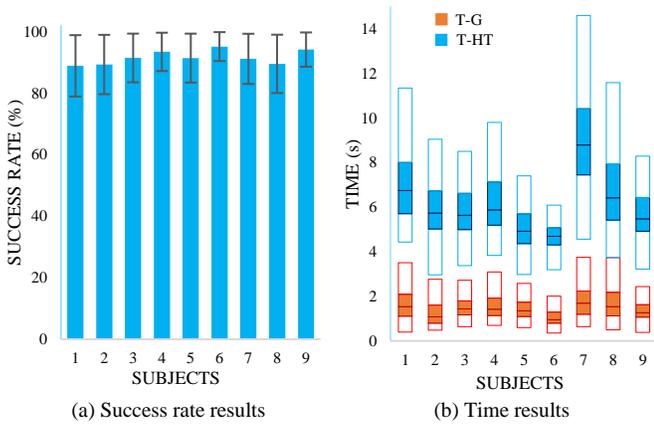

(a) Success rate results  (b) Time results
Fig. 8 Results of different healthy subjects

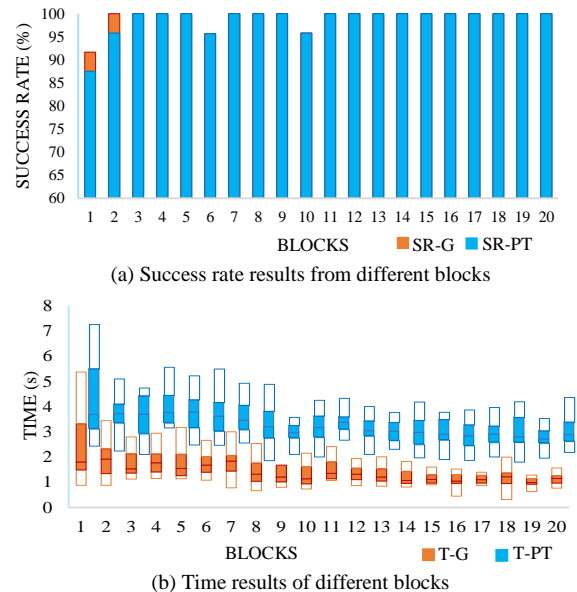

(a) Success rate results from different blocks

(b) Time results of different blocks
Fig. 9. Results from different blocks of the patient.

the index and thumb fingers. The objects grasped by the *Cylindrical, Spherical, and Tripod* were easier to transfer, with a 95%-plus success rate. All healthy subjects thought it was challenging to grasp and transfer the bead, and only 66% of cases were successful, as shown in Figure 7(c). To grasp a bead stably requires good balance in the control of position and strength to prevent the bead from being squeezed out or dropped. The transdermal patch was the easiest to transfer, with a 98%-plus success rate. The analysis of the time showed (shown in Figure 7(b)) that the objects grasped by the *Lateral* needed the most time (around 6.2 s) to transfer as the *Lateral* needed more time to pre-shape and grasp, and there were no statistical differences on other grasp types, about 5.5 to 5.8 s. The test for T-G on different grasp types found no significant differences ($p = 0.7$).

Significant differences in the healthy subjects remained in the time spent transferring since everyone's ability to operate a new system is different. As shown in Figure 8, the sixth subject had the highest success rate in transferring an object, around 95%, and also spent the least time, about 4.7 s. The first subject had the lowest success rate, around 88%. The seventh subject spent the most time completing the transfer task, around 8.8 s, and also spent the most time switching the grasp type.

The result of the patient with the longer test (twenty blocks) showed a significant improvement in performance throughout the test. Figure 9(a) shows that he accomplished the task in only 87% of trials in the first block and 96% in the second block, but this number remained at 100% after the eleventh block. The time spent to accomplish tasks, shown in Figure 9(b), decreased significantly as the number of trials increased, and it just needed about 1.1 s to switch the target and around 2.8 s to grasp the object in the last five blocks.

## V. DISCUSSIONS

A new hybrid prosthetic control system (i-MYO) is proposed in the paper. With the help of the i-MYO system, users can quickly, easily, stably, and robustly control the prosthetic hand to manipulate daily life objects using a variety of grasp types. The new grasp-type switcher can switch quicker among a larger quantity of grasp types than traditional methods, such as Co-Contraction and coded switching methods [2]. The users are also more relaxed and less fatigued since they only need to glance at a GazeButton on the grasp-type panel to switch a grasp type. In the past, the user was usually required to repeatedly contract the forearm muscles, which was tedious and tiring. The result of the i-MYO system was very stable because the i-MYO system provides intuitive real-time feedback of the grasp-type results to allow the users to make corrections in time. The method is not affected by various confounding factors like EMG (e.g., electrode deviation, sweating, tension, temperature changes, muscle fatigue).

According to the modular design idea, the grasp-type interface (i-GSI) of the i-MYO system is designed as a completely independent module. The module can run independently on a mobile headset device (e.g., HoloLens 2) since its computational burden is small. The module does not need to receive signals from other devices. The grasp-type result in this module can be transmitted to the existing prosthetic system through wireless communication (e.g., WIFI and Bluetooth). Most commercial dexterous prosthetic hands are manufactured with built-in wireless communication to receive a grasp type from a mobile phone APP. The new switching interface can be easily added to its program frame to replace the mobile APP (e.g., an i-Limb prosthetic hand and communicate with Bluetooth [45]).

It is found that during the operation of the i-MYO system, switching a grasp type and manipulating the hand is not strictly controlled in sequence, and it may be a parallel mechanism. For example, the subject will observe the scene first after switching the grasp type in most cases and then manipulate the hand to reach an object and trigger the EMG at the same time to pre-shape the hand when ensuring the safety of the prosthetic hand operation. However, after many trials and proficiency, some subjects boldly raised and pre-shaped the robot hand first instead of observing the scene when the sight was still on the grasp-type panel (the target GazeButton had been triggered).

Although the i-GSI has achieved good results in the experiment, failed trials still exist. For example, when the robot hand reaches the object's vicinity, the grasp type will occasionally be mistakenly switched to another grasp type. The main reason is that the HoloLens 2 has a limited holographic field of view (in the vertical direction), and the grasp-type panel cannot be completely out from the common sight area. In this way, an error may happen when the grasp-type panel overlaps with the supervision area where the user supervises the robot hand. However, devices with a wider field of view and even an infinite field of view are rapidly developing (e.g., HoloLens 3[46]), which may solve this problem.

Furthermore, with the development of mechatronics technology, there have been dexterous prosthetic hands equipped with active wrists (e.g., i-Limb® Quantum[45]), which poses a higher challenge to the traditional man-machine interface. Although this study only evaluated the performance of the i-MYO system in controlling the fingers, the proposed system is also applicable to control an active wrist with multiple degrees of freedom, such as pronation/supination and flexion/extension. A two-DOF wrist has been integrated into our prosthetic hand, and some simple tests have been conducted. In the future, we will carry out wrist-hand combination paradigm experiments to evaluate the performance of the new method in higher-level prosthetics and more complex tasks.

## VI. Conclusions

A new hybrid prosthetic control system (i-MYO) based on eye-tracking, augmented reality feedback, and EMG is proposed in this study. Nine healthy subjects and one patient tested the operation of the i-MYO. Our results show that a novice can quickly master the operation of the new control system after a few trials. With the new control system, the user can easily switch between six grasp types by glancing at a GazeButton and intuitively control the prosthetic hand to transfer the daily life objects together with a simple EMG interface. Our endeavor on a primary clinical trial also shows that the new method can also be adapted to a patient with extremely poor EMG. The new method largely lessens the difficulty for users to control a dexterous prosthetic hand comfortably.

## VII. ACKNOWLEDGMENTS